\documentclass[sigconf]{acmart}

\usepackage{multirow}
\usepackage{float}

\AtBeginDocument{%
  }

\copyrightyear{2026}
\acmYear{2026}
\setcopyright{cc}
\setcctype{by}
\acmConference[SIGSPATIAL '26]{The 34th ACM International Conference on Advances in Geographic Information Systems}{November 03--06, 2026}{Riverside, CA, USA}
\acmBooktitle{The 34th ACM International Conference on Advances in Geographic Information Systems (SIGSPATIAL '26), November 03--06, 2026, Riverside, CA, USA}
\acmDOI{10.1145/3841645.3842989}
\acmISBN{979-8-4007-2950-8/2026/11}

\begin{document}
\raggedbottom

\title[Do Satellites See Commuters? A Critical Benchmark of Vision Foundation Models]{Do Satellites See Commuters? A Critical Benchmark of Vision Foundation Models for Generative Urban Flow Networks}

\author{Ashiq Shukoor Iqbal}
\affiliation{%
  \institution{The University of New South Wales}
  \city{Sydney}
  \state{NSW}
  \country{Australia}
}
\email{ashiq\_shukoor.iqbal@student.unsw.edu.au}

\author{Wilson Wongso}
\affiliation{%
  \institution{The University of New South Wales}
  \city{Sydney}
  \state{NSW}
  \country{Australia}
}
\email{w.wongso@unsw.edu.au}

\author{Flora Salim}
\affiliation{%
  \institution{The University of New South Wales}
  \city{Sydney}
  \state{NSW}
  \country{Australia}
}
\email{flora.salim@unsw.edu.au}

% This shortens the header on subsequent pages to prevent text overlap
\renewcommand{\shortauthors}{Iqbal, Wongso, and Salim}

\begin{abstract}
Satellite foundation models offer a globally available alternative to census data for commuting origin-destination (OD) generation, yet no study has systematically compared encoder paradigms within a single downstream pipeline. We ablate four satellite vision encoders: language-supervised (RemoteCLIP), self-supervised (DINOv3), and geographically grounded (SatCLIP, AlphaEarth) within an identical WeDAN graph diffusion framework across 1,925 US counties, 325 UK districts, and 14 global cities under five random seeds. Three main findings emerge. First, language-supervised features achieve the strongest in-distribution performance (RemoteCLIP CPC 0.602), while geographically grounded encoders transfer more reliably zero-shot: AlphaEarth improves CPC by 33\% over RemoteCLIP on UK districts. Second, pretraining corpus scale alone is insufficient: DINOv3, trained on a substantially larger satellite corpus, underperforms RemoteCLIP by 0.091 CPC in-distribution and collapses to CPC 0.022 globally. Third, no encoder transfers usefully to global cities (best CPC 0.122 for RemoteCLIP, 0.022 for DINOv3), confirming cross-continental OD generation remains an open problem. We additionally clarify the semantics of the census noise parameter~$\eta$, whose ordering reverses under cross-continental evaluation, a distinction critical to correctly interpreting prior results. Our code is available at: \url{https://github.com/cruiseresearchgroup/satellites-see-commuters}.
\end{abstract}




\maketitle
\section{Introduction}

Modeling urban mobility patterns via Origin-Destination (OD) matrices represents a cornerstone of contemporary civil infrastructure planning, transportation system optimization, and macro-level spatiotemporal analysis. Historically, the extraction of these granular travel flow networks has depended heavily on localized administrative census datasets, household travel surveys, or aggregated signaling logs sourced from proprietary mobile network operators. However, these traditional collection methods suffer from severe financial and operational bottlenecks: comprehensive census registries are typically updated only at decadal intervals, household surveys scale poorly due to low response frequencies, and telecommunication logs are strictly constrained by privacy regulations and commercial siloing. This pervasive data gap is particularly acute within medium-sized municipalities in emerging economies and rapidly transforming urban zones, where the absence of baseline demographic records leaves municipal planners without actionable structural insights.

To circumvent these multi-source data dependencies, recent deep generative architectures have reframed the problem of mobility network synthesis, modeling OD generation as a conditional graph denoising task. Architectures like GlODGen~\cite{glodgen} completely eliminate the reliance on downstream tabular census parameters by utilizing high-resolution satellite imagery and globally uniform WorldPop statistics as their exclusive input modalities. By processing region-level satellite tiles through RemoteCLIP~\cite{remoteclip}, a vision-language foundation model pre-trained on remote sensing imagery and natural language captions, and conditioning the underlying WeDAN graph diffusion pipeline~\cite{wedan} on the resulting visual features, this framework demonstrates that satellite-derived signals are highly expressive of underlying human commuting patterns. Under in-distribution settings, this satellite-conditioned configuration achieves a Common Part of Commuters (CPC) score of 0.623 across 1,925 US counties, recovering 98.3\% of full-census WeDAN performance (CPC 0.623 vs. 0.634). Our replication of GlODGen under identical conditions yields $\text{CPC} = 0.602$, confirming the encoder's signal while establishing the ablation baseline.

Despite these strong results, existing frameworks treat the underlying satellite vision encoder backbone, RemoteCLIP, as an architectural primitive. Since the initial formulation of satellite-conditioned graph diffusion, the geospatial foundation model landscape has expanded significantly, yielding distinct self-supervised paradigms optimized via fundamentally distinct pretraining objectives. For example, self-distillation networks such as DINOv3~\cite{dinov3} leverage massive, unlabeled image repositories to capture fine-grained pixel textures and geometric boundaries. Coordinate-contrastive models like SatCLIP~\cite{satclip} map localized visual patches directly to universal geographic coordinates to enforce explicit spatial grounding. Finally, multi-sensor frameworks such as AlphaEarth~\cite{alphaearth} combine diverse multi-spectral, radar, and elevation datastreams into highly compressed digital signatures. The current literature lacks a systematic comparison of vision encoders within a fixed generative graph pipeline, while these encoders differ in objective, dimension, and preprocessing. Consequently, it remains an open question whether scene-level text-aligned features via RemoteCLIP are a strict requirement for modeling human mobility, or if self-supervised structural representations can provide equivalent or superior downstream generalizability.

This paper directly addresses this research gap by presenting a systematic empirical evaluation of geospatial vision foundation models embedded within generative graph diffusion architectures. Keeping the same downstream WeDAN graph transformer pipeline, we evaluate the representational fidelity of four vision encoder paradigms across a multi-scale benchmarking pipeline. Our experimental framework evaluates in-distribution performance using 1,925 county-level networks within the United States, followed by zero-shot cross-continental transferability testing across 325 Local Authority Districts in the United Kingdom and 14 morphologically distinct global metropolitan cities. Every configuration is evaluated across five independent random seeds, using paired $t$-tests over matched seeds to verify the statistical reliability of observed performance changes. Our main contributions are summarized as follows:
\begin{itemize}
    \item We conduct the first controlled empirical ablation isolating the impact of visual representation spaces on generative OD graph diffusion, evaluating text-aligned models (RemoteCLIP), self-distillation architectures (DINOv3), coordinate-contrastive models (SatCLIP), and multi-sensor physical embeddings (AlphaEarth).
    \item We formally clarify the semantics of the census noise parameter ($\eta$) in the WeDAN architecture, demonstrating through architectural analysis that it measures census data availability rather than satellite feature noise, which establishes the correct operating point for cross-continental evaluation.
    \item We analyze a representational tradeoff across vision encoders that differ in pretraining corpus scale, objective, architecture, and preprocessing, highlighting that larger-scale pretraining does not necessarily translate to better representations for spatially aggregated tasks (DINOv3 achieving $\text{CPC} = 0.511$ vs. RemoteCLIP achieving $\text{CPC} = 0.602$, $p < 0.001$).
    \item We demonstrate that for cross-continental zero-shot transfer, geographically grounded encoders significantly outperform language-aligned features on the UK evaluation dataset, with multi-sensor physical embeddings (AlphaEarth) achieving a performance margin of 0.127 CPC over RemoteCLIP ($p = 0.001$, paired $t$-test).
\end{itemize}

The remainder of this paper is structured as follows. Section 2 reviews the historical and contemporary literature across mobility modeling, geospatial foundation models, and domain generalization boundaries. Section 3 formalizes the problem formulation, graph diffusion mechanics, and the theoretical interpretation of the census noise parameter. Section 4 outlines our multi-scale dataset profiles, geospatial encoder adaptation mechanisms, and baseline implementation details. Section 5 presents our comprehensive empirical findings, representational analyses, and cross-continental scaling anomalies. Section 6 concludes with practical recommendations and future directions.

\section{Background and Related Work}

\subsection{Deep Learning for Human Mobility}
The mathematical modeling of Origin-Destination (OD) flow generation has shifted from rigid physics-inspired spatial interaction models to flexible, data-driven deep architectures. Traditional spatial interaction baselines are anchored by the Gravity model~\cite{gravity}, which operates on an analogy to Newtonian mechanics where commuting flows are directly proportional to regional population masses and inversely proportional to a power function of geographic distance. Similarly, the Radiation model~\cite{radiation} eliminates parameter tuning by calculating flow systems based on job vacancy distributions and surrounding population densities. While these classical systems provide basic spatial priors, they are entirely macroscopic, with classic Gravity approaches limited to a lower performance baseline ($\text{CPC} \approx 0.32$). They consistently fail to capture the highly localized, non-linear functional zoning dependencies, such as specialized commercial corridors or residential pockets, that govern modern metropolitan transit.

To bypass these strict parametric assumptions, researchers transitioned to pair-wise deep learning paradigms. Architectures like DeepGravity~\cite{deepgravity} and the Geo-contextual Multitask Embedding Learner (GMEL)~\cite{gmel} deploy deep feed-forward networks to map localized neighborhood land-use features, computing flow probabilities across decoupled origin and destination vector pairs. However, optimizing these independent pairs ignores the systemic, network-wide dynamics that dictate physical human transit; global OD marginal sums are conservation constraints that pair-wise models cannot enforce. 

This critical systemic limitation forced the adoption of Graph Neural Networks (GNNs). Frameworks such as ODCRN~\cite{odcrn} map urban environments as cohesive topological structures, tracking regional transit corridors and capturing complex spatial structural constraints through message-passing mechanics.

Most recently, generative graph diffusion paradigms have redefined performance benchmarks. The WeDAN architecture~\cite{wedan} treats the generation of complete OD matrices as a parameterized reverse denoising process, leveraging continuous Gaussian noise corruption to capture both fine-grained local anomalies and macro-level spatial structures, establishing a new state-of-the-art baseline ($\text{CPC} \approx 0.59$ on the full 3,233-area LODES benchmark).

Building directly upon this diffusion framework, GlODGen~\cite{glodgen} introduced satellite-conditioned node embeddings to eliminate the downstream network's historical reliance on highly localized, expensive, and non-generalizable tabular census registries. However, GlODGen permanently fixes RemoteCLIP as its vision backbone. Our work directly challenges this assumption, systematically evaluating whether alternative vision encoders yield stronger downstream predictive abilities and cross-continental robustness.%alternative spatial pretraining objectives yield superior alignment and cross-continental resilience.

\subsection{Spatial Vision Foundation Models}
The rapid expansion of self-supervised computer vision has yielded highly specialized remote sensing foundation models. Rather than sharing a unified representational space, these models are optimized under fundamentally different pretraining paradigms, resulting in distinct representational properties. Table~\ref{tab:encoder_comparison} outlines the core architectural profiles evaluated within our benchmark.

\begin{table*}[t]
\centering
\small
\caption{Architectural and Pretraining Comparison of Evaluated Spatial Vision Foundation Models}
\label{tab:encoder_comparison}
\begin{tabular}{lcccc}
\hline
\textbf{Model} & \textbf{Architecture} & \textbf{Dim} & \textbf{Pretraining} & \textbf{Supervision} \\ \hline
RemoteCLIP & ViT-L/14 & 1024 & RS image-text pairs & Contrastive (VLM) \\
DINOv3 & ViT-L/16 & 1024 & 493M satellite images & Self-distillation \\
SatCLIP & ViT-S/16 & 384 & GPS-matched patches & Coordinate-contrastive \\
AlphaEarth & Multi-sensor GEE fusion & 64 & 6-sensor fusion & Self-supervised (GEE) \\ \hline
\end{tabular}
\end{table*}

\begin{itemize}
    \item \textbf{Vision-Language Alignment:} RemoteCLIP~\cite{remoteclip} utilizes a massive contrastive language-image pretraining (CLIP) objective. By aligning remote sensing imagery directly with corresponding descriptive natural language tokens, the model embeds high-level anthropogenic geography, functional zoning concepts, and socioeconomic semantics directly into its 1024-dimensional latent space.
    \item \textbf{Self-Supervised Patch Discrimination:} DINOv3~\cite{dinov3} optimizes a vision transformer backbone using a teacher-student self-distillation objective over a corpus of 493 million satellite images. This framework forces the network to specialize in localized patch discrimination, prioritizing high-frequency geometric features, building boundaries, and sharp surface textures over abstract semantic meanings.
    \item \textbf{Coordinate-Contrastive Grounding:} SatCLIP~\cite{satclip} discards language completely, deploying a coordinate-contrastive objective that forces the visual transformer to map satellite patches directly to their exact global GPS coordinates. The resulting 384-dimensional latent space is implicitly structured by continuous geographic distance, capturing spatial variations across the Earth's surface.
    \item \textbf{Multisensor Physical Grounding:} AlphaEarth~\cite{alphaearth} shifts away from heavy visual transformers entirely, deploying a multi-datastream network natively hosted within the Google Earth Engine (GEE) cloud architecture. By assimilating six distinct sensor modalities across optical, radar, and elevation datastreams, it extracts explicit surface moisture, spectral indices, and vegetation canopy structure into a highly compact 64-dimensional feature vector.
\end{itemize}

\subsection{The Challenge of Zero-Shot Cross-Continental Transfer}
Deploying predictive mobility models across different domains and geographical regions remains an open challenge. Domain generalization for GNNs has been explored through distribution shift minimization~\cite{zhu2021shift}, though these methods target node classification rather than generative flow estimation. Concurrently, the remote sensing community has established benchmarks like BigEarthNet~\cite{bigearthnet} to test the cross-domain transferability of representation layers across highly heterogeneous multi-spectral landscapes spanning multiple European nations.

Despite these advancements in visual classification and domain-invariant graph mining, the problem of direct cross-continental transfer in generative human mobility remains structurally unaddressed. Predicting commuting volumes across completely different continents introduces severe, compounding covariates. Local architectural forms, street network configurations, regional population densities, transit infrastructure designs, and socio-environmental zoning behaviors shift radically between North American developments and European historic grids. 

When a model is trained exclusively on domestic distributions, its conditioning layers easily overfit to regional features. To our knowledge, no prior work evaluates satellite-conditioned OD generation under zero-shot cross-continental conditions: our two-stage transfer evaluation (US $\to$ UK and US $\to$ Global) serves as the first systematic benchmark to uncover how these shifting visual and structural topologies impact generative graph diffusion networks.

\subsection{Evaluation Protocols for OD Generation}
Validating the systemic accuracy of a generated graph adjacency matrix requires mathematical evaluation protocols that can isolate relative spatial routing logic from absolute numerical scaling behavior. While RMSE and MAE are widely used in traffic forecasting and vehicle routing, these metrics operate purely on absolute arithmetic differences. In zero-shot cross-continental settings, absolute magnitude estimation is highly fragile: subtle distribution shifts in a foundation model's latent conditioning vector can cause a downstream graph transformer to output runaway volume scales, completely disrupting absolute error metrics even if the directional routing logic remains accurate.

To establish a scale-invariant validation framework, modern origin-destination literature prioritizes the Common Part of Commuters (CPC) metric~\cite{lenormand2012universal}. CPC calculates the normalized overlap between the full ground-truth and generated OD distributions as a global coefficient over all origin-destination pairs. By explicitly penalizing misallocated spatial connections while remaining substantially more robust to absolute magnitude scaling than regression-based metrics, CPC provides the stable, topology-focused signal required to evaluate true cross-continental alignment reliability.

\section{Problem Formulation \& Architecture}

\subsection{OD Generation as Conditional Graph Diffusion}
We model a city as a directed graph $G = (V, E)$, where nodes $v \in V$ represent geographic regions, such as census tracts or administrative zones, and edges $e_{ij} \in E$ represent the commuting flow $F_{ij}$ from origin $i$ to destination $j$. The objective of the origin-destination (OD) generation task is to construct the complete flow adjacency matrix $\mathbf{F} \in \mathbb{R}^{|V| \times |V|}$. This matrix is generated by an architecture trained as a Denoising Diffusion Probabilistic Model (DDPM)~\cite{ddpm} and sampled at inference via a Denoising Diffusion Implicit Model (DDIM)~\cite{ddim}.

The forward diffusion process systematically adds Gaussian noise to the true flow matrix $\mathbf{F}_0$ over $T$ discrete timesteps according to a predefined variance schedule $\beta_1, \beta_2, \dots, \beta_T$:
\begin{equation}
q(\mathbf{F}_t \mid \mathbf{F}_{t-1}) = \mathcal{N}(\mathbf{F}_t; \sqrt{1 - \beta_t}\mathbf{F}_{t-1}, \beta_t\mathbf{I})
\end{equation}
Using the notation $\alpha_t = 1 - \beta_t$ and $\bar{\alpha}_t = \prod_{i=1}^t \alpha_i$, we can express the marginalized distribution at any arbitrary timestep $t$ directly as:
\begin{equation}
q(\mathbf{F}_t \mid \mathbf{F}_0) = \mathcal{N}(\mathbf{F}_t; \sqrt{\bar{\alpha}_t}\mathbf{F}_0, (1 - \bar{\alpha}_t)\mathbf{I})
\end{equation}

Following the WeDAN architecture~\cite{wedan}, the reverse denoising process employs a GraphTransformer network~\cite{graphtransformer} to iteratively remove noise from the continuous flow matrix over a series of discrete timesteps. Crucially, this generation process is conditioned on node-level feature vectors, ensuring that the unique socio-environmental and structural characteristics of each geographic region directly guide the predicted downstream mobility flows.

\subsection{Satellite-Conditioned Node Features}
In traditional mobility frameworks, the node feature vector relies entirely on localized, tabular census statistics. In our satellite-conditioned paradigm, the base feature vector for a given region $r$ is defined as:
\begin{equation}
\mathbf{x}_r = [\varphi(I_r) \, \| \, \log(1 + \mathbf{p}_r)] \in \mathbb{R}^{d+2}
\end{equation}
where $\varphi(I_r) \in \mathbb{R}^d$ represents the embedding vector produced by a spatial vision foundation model applied to the satellite imagery of the region, and $\mathbf{p}_r \in \mathbb{R}^2$ represents globally available WorldPop statistics covering total population and land area. To ensure highly uniform spatial sampling across irregular administrative boundaries, each geographic region is partitioned using H3 resolution 9 hexagonal cells. The selected vision encoder processes each hexagonal tile independently, and the resulting individual embeddings are mean-pooled across all tiles contained within the region boundary to generate the final visual representation $\varphi(I_r)$.

During model training, the final conditioning vector fuses the satellite representation $\mathbf{x}_r$ with a noise-controlled census vector $\tilde{\mathbf{a}}_r$:
\begin{equation}
\tilde{\mathbf{a}}_r = \mathbf{a}_r \cdot \eta + \boldsymbol{\varepsilon} \cdot (1-\eta)
\end{equation}
where $\mathbf{a}_r \in \mathbb{R}^{97}$ is the true localized census data vector and $\boldsymbol{\varepsilon} \sim \mathcal{N}(\mathbf{0}, \mathbf{I})$ represents standard Gaussian noise. 

The full node vector fed to the GraphTransformer is then constructed as:
\begin{equation}
\mathbf{n}_r = [\mathbf{x}_r \,\|\, \tilde{\mathbf{a}}_r \,\|\, \hat{\mathbf{a}}_r] \in \mathbb{R}^{(d+2)+97+97}
\end{equation}
where $\hat{\mathbf{a}}_r$ is a DenoisingTransformer prediction of census from satellite features. In all experiments, setting the configuration parameter \texttt{if-ImgAttrAug=0} zeros the predicted vector $\hat{\mathbf{a}}_r$, meaning the effective input simplifies to: 
\begin{equation}
\mathbf{n}_r = [\mathbf{x}_r \,\|\, \tilde{\mathbf{a}}_r \,\|\, \mathbf{0}].
\end{equation}
This structural formulation explains why the downstream network input dimension scales directly with the choice of spatial foundation model, varying based on the underlying dimensionality $d$ of the vision encoder.

\begin{figure*}[tbp]
  \centering
  \includegraphics[width=\textwidth]{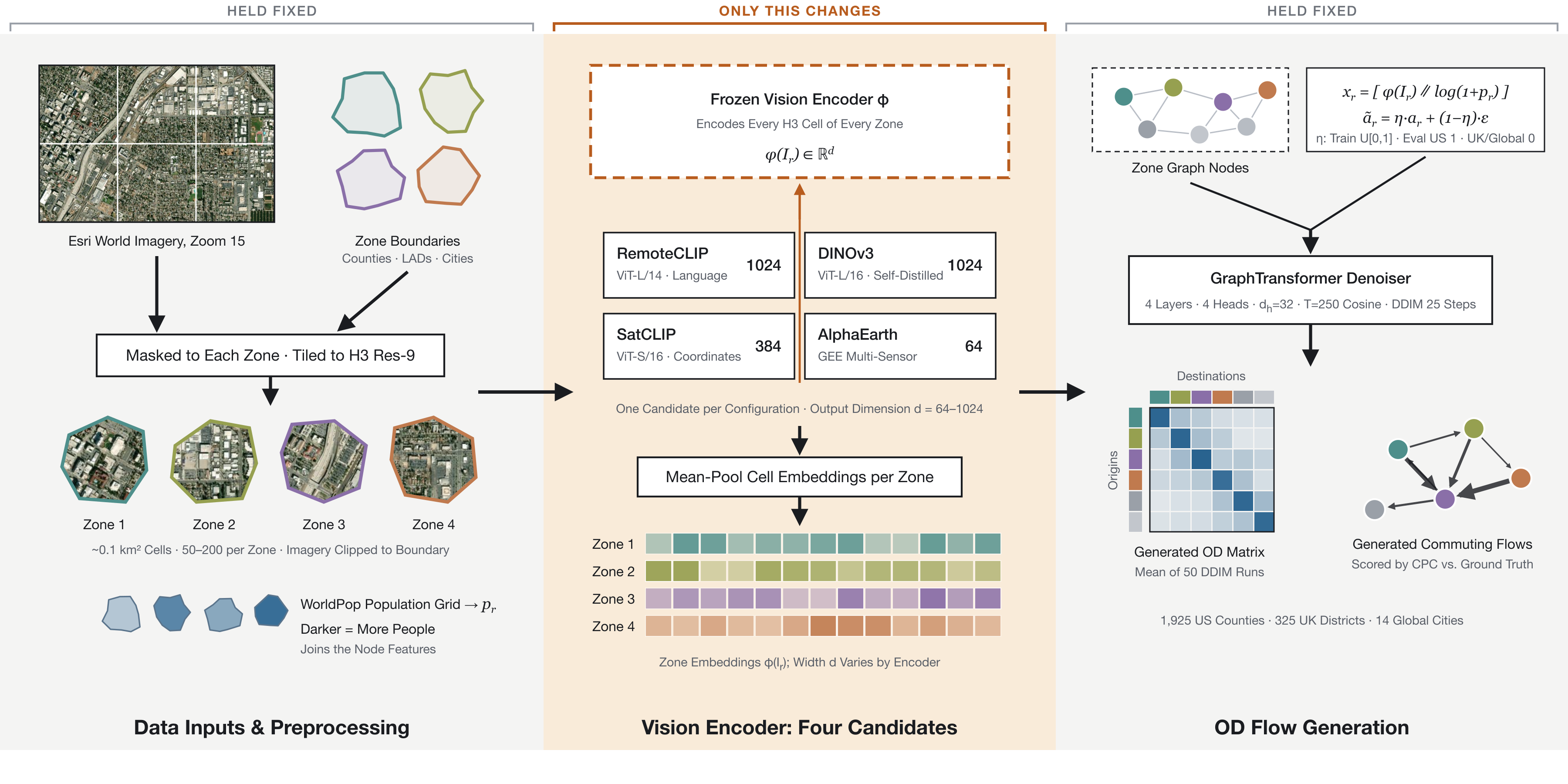}
  \caption{GlODGen pipeline. Satellite imagery and population data are preprocessed via H3 hexagonal tiling (resolution 9, ${\sim}0.1\,\text{km}^2$ per cell) with per-tile mean pooling. A spatial vision encoder extracts zone-level embeddings conditioning the WeDAN GraphTransformer diffusion model, which generates the OD flow matrix via DDIM sampling.}
  \label{fig:pipeline}
\end{figure*}

\subsection{GraphTransformer Denoising Mechanics and Training Loss}
The GraphTransformer network updates the node states and processes edge relationships within the flow graph. Given the concatenated matrix of node feature vectors $\mathbf{H}^{(0)} = [\mathbf{n}_1, \mathbf{n}_2, \dots, \mathbf{n}_{|V|}]^T$, each transformer layer $l$ applies a multi-head attention mechanism. For a specific attention head $k$, the query, key, and value matrices are projected using learned parameter weights:
\begin{equation}
\mathbf{Q}_k = \mathbf{H}^{(l)}\mathbf{W}_k^Q, \quad \mathbf{K}_k = \mathbf{H}^{(l)}\mathbf{W}_k^K, \quad \mathbf{V}_k = \mathbf{H}^{(l)}\mathbf{W}_k^V
\end{equation}
The spatial interaction between node $i$ and node $j$ is computed by taking the scaled dot product of their respective projections, incorporating a temporal step embedding $\mathbf{t}$ to track the diffusion timeline:
\begin{equation}
\mathbf{A}_k(i, j) = \frac{(\mathbf{Q}_k)_i (\mathbf{K}_k)_j^T}{\sqrt{d_{head}}} + \psi_k([\mathbf{F}_t]_{ij}, \mathbf{t})
\end{equation}
where $\psi_k$ represents an edge projection network that transforms the noisy flow state $[\mathbf{F}_t]_{ij}$ into a structural bias modifier for the attention matrix. The updated node hidden representations are then compiled across all $K$ heads using a linear projection layer and a feed-forward network:
\begin{equation}
\mathbf{H}^{(l+1)} = \text{FFN}\left(\left[\text{Softmax}(\mathbf{A}_1)\mathbf{V}_1 \,\|\, \dots \,\|\, \text{Softmax}(\mathbf{A}_K)\mathbf{V}_K\right]\mathbf{W}^O\right)
\end{equation}

The network parameter optimization relies on a simplified mean squared error objective function. The loss function forces the GraphTransformer to isolate and predict the precise Gaussian noise vector $\boldsymbol{\epsilon}$ injected into the flow matrix at any given diffusion step $t$:
\begin{equation}
\mathcal{L}_{\text{simple}}(\theta) = \mathbb{E}_{t \sim \mathcal{U}\{1,T\},\, \mathbf{F}_0 \sim q,\, \boldsymbol{\epsilon} \sim \mathcal{N}(\mathbf{0},\mathbf{I})}\left[ \left\| \boldsymbol{\epsilon} - \boldsymbol{\epsilon}_\theta(\mathbf{F}_t, \mathbf{H}^{(0)}, \mathbf{t}) \right\|_F^2 \right]
\end{equation}
By minimizing this objective across all discrete timesteps, $\theta$ learns to reverse the forward corruption process. At inference, the trained GraphTransformer is deployed within DDIM sampling~\cite{ddim}, which generates structured flow matrices via a deterministic 25-step trajectory, yielding a 10$\times$ reduction from the $T{=}250$ training steps. To reduce stochastic variance, the final output for each city is averaged over 50 independent DDIM runs. The node-level conditioning $\mathbf{H}^{(0)}$, fixed at initialization, ensures that all generated flows remain anchored to the satellite and demographic representations of each geographic zone.

\subsection{Semantics of the Census Noise Parameter}
Our architectural analysis indicates that the census noise parameter ($\eta$) is best understood as governing the assumed \emph{availability} of true census attributes, rather than acting as a noise-injection regularizer on satellite features. During training, $\eta$ is resampled uniformly at random for each city in each batch, so the model observes the full range of census availability, from clean attributes through to a purely noisy channel. It is never exposed to an all-zero census vector, however: that would require $\mathbf{a}_r = \mathbf{0}$ and $\eta = 1$ to coincide, which does not occur under continuous sampling of $\eta$. This insight reveals two distinct operational paradigms depending on the evaluation context:
\begin{itemize}
    \item \textbf{In-Distribution Evaluation:} When true localized census data $\mathbf{a}_r$ is fully available, setting $\eta = 1$ provides the model with clean, multi-modal conditioning. Setting $\eta = 0$ replaces the census data entirely with Gaussian noise, forcing the architecture to rely solely on the satellite representation $\mathbf{x}_r$ to generate flows.
    \item \textbf{Zero-Shot Cross-Continental Transfer:} In target regions where local census datasets are missing or structurally incompatible, true census data is unavailable; supplying $\mathbf{a}_r = \mathbf{0}$ as a surrogate alongside a setting of $\eta = 1$ forces an all-zero vector into the network, yielding $\tilde{\mathbf{a}}_r = \mathbf{0}$. Because $\eta$ is randomized throughout training, the model has repeatedly encountered a noise-dominated census channel, but never an all-zero one. That substitution is therefore genuinely out of distribution, whereas evaluating at $\eta = 0$ is not: it reproduces a conditioning regime the model was trained under. This asymmetry, rather than a generic distribution shift, is what separates the two substitutions.
\end{itemize}
Accordingly, we report all US in-distribution results at $\eta = 1$ and all UK and Global zero-shot results at $\eta = 0$, and we treat these as the declared operating points for every headline comparison. The encoders do not, however, respond uniformly to $\eta$ on the UK data (Figure~\ref{fig:cpc_eta}): AlphaEarth degrades sharply as $\eta \to 1$ (0.515 to 0.249) while DINOv3 remains essentially flat. We do not claim a single mechanism for that spread.

\section{Experimental Setup}

\subsection{Datasets and Evaluation Metrics}
Our empirical evaluation spans three distinct geographic scales to rigorously stress-test both in-distribution performance profiles and zero-shot cross-continental generalization capabilities:
\begin{itemize}
    \item \textbf{US In-Distribution:} This dataset serves as the foundational training and localized testing corpus, consisting of 1,925 counties retained from the full LODES pool following GlODGen's data filtering to include only areas with a complete OD matrix and satellite feature data. The ground-truth commuting configurations are sourced from the Longitudinal Employer-Household Dynamics Origin-Destination Employment Statistics (LODES) 2018 Census Bureau repository. For every evaluation seed, this dataset is partitioned into an 80/10/10 random split dedicated to model training, internal hyperparameter validation, and in-distribution testing, respectively.
    \item \textbf{UK Zero-Shot Transfer:} To evaluate strict zero-shot cross-continental generalizability, we introduce an evaluation target incorporating 325 Local Authority Districts within the United Kingdom. Ground-truth cross-zonal commuting flows are taken from the UK benchmark released with GlODGen~\cite{glodgen}, which derives population-level flows between census units from Office for National Statistics (ONS) registries. Under this cross-domain evaluation paradigm, the downstream generative model is trained exclusively on North American US urban areas, completely omitting any target-domain fine-tuning or structural adaptation before inference.
    \item \textbf{Global Zero-Shot Evaluation:} To assess unconstrained scalability across highly heterogeneous urban forms, this evaluation maps 14 morphologically diverse international metropolitan areas outside the US and UK core regions: Baoding, Beijing, Chengdu, Guangzhou, London, Namibia, Paris, Rio de Janeiro, Senegal, Shanghai, Shenzhen, Sydney, Tangshan, and Tokyo. The same US-trained architecture is frozen and applied directly to these international regions without weight optimization, representing the more distant of two zero-shot targets: US $\to$ Global. Ground-truth flow baselines for these global metropolises are aggregated from localized household travel surveys and anonymous mobile network signaling registries~\cite{glodgen}.
\end{itemize}

These three targets differ in reference year, zoning system, and collection method, and the satellite composites are not date-matched to any of them; cross-region differences therefore reflect differences in reference data as well as in urban morphology.

The primary metric utilized to quantify spatial allocation accuracy across all experimental boundaries is the Common Part of Commuters (CPC). Let $\mathbf{F} \in \mathbb{R}^{|V| \times |V|}$ represent the ground-truth origin-destination flow matrix, and let $\hat{\mathbf{F}} \in \mathbb{R}^{|V| \times |V|}$ represent the corresponding matrix generated by the diffusion network. The CPC score is mathematically formalized as:
\begin{equation}
\text{CPC}(\mathbf{F}, \hat{\mathbf{F}}) = \frac{2 \sum_{i=1}^{|V|} \sum_{j=1}^{|V|} \min(F_{ij}, \hat{F}_{ij})}{\sum_{i=1}^{|V|} \sum_{j=1}^{|V|} F_{ij} + \sum_{i=1}^{|V|} \sum_{j=1}^{|V|} \hat{F}_{ij}}
\end{equation}

The CPC metric calculates the normalized overlap coefficient between the true and predicted mobility distributions, where $\text{CPC} \in [0, 1]$. A score of 1 indicates perfect structural reconstruction, whereas a score of 0 denotes complete spatial disconnection. In addition to CPC, standard error scales including Root Mean Squared Error (RMSE), normalized Root Mean Squared Error (NRMSE), and Mean Absolute Error (MAE) are tracked to assess numeric scaling behavior.

\subsection{Encoder Adaptation and Feature Extraction}
High-resolution satellite imagery tiles are retrieved from the Esri World Imagery repository at zoom level 15 using the automated cloud pipelines of Google Earth Engine. To ensure highly uniform visual sampling across irregular administrative and political boundaries, each target geographic region is partitioned using an Uber H3 spatial index at resolution 9, generating uniform hexagonal cells with an average area of approximately 0.1 square kilometers. 

Each spatial foundation model processes the individual hexagonal tiles falling within a region independently. The resulting collection of high-dimensional latent tile embeddings is subsequently mean-pooled across the region administrative boundary to construct a single, unified visual representation vector $\varphi(I_r)$. The four underlying vision encoder adapt their internal feature extraction layers as follows:
\begin{itemize}
    \item \textbf{RemoteCLIP:} We deploy the ViT-L/14 visual backbone to extract dense 1024-dimensional visual embeddings. This representation benefits directly from pre-aligned language supervision, capturing high-level anthropogenic land-use patterns.
    \item \textbf{DINOv3:} We use the satellite-pretrained \texttt{dinov3-vitl16}\allowbreak\texttt{-pretrain-sat493m} checkpoint (ViT-L/16) to obtain 1024-dimensional embeddings. These features are trained via self-supervised teacher-student distillation across a corpus of 493 million unannotated satellite images, rendering the latent features highly sensitive to micro-level surface geometry and texture.
    
   \item \textbf{SatCLIP:} This paradigm utilizes a ViT-S/16 backbone to generate 384-dimensional coordinate-aligned feature maps. The SatCLIP checkpoint's patch embedding projection is pretrained on 13-band Sentinel-2 imagery~\cite{sentinel2}, producing weights of shape $[384, 13, 16, 16]$. To adapt to 3-channel RGB tiles, we average these weights across the 13 spectral bands to obtain a single-channel kernel $[384, 1, 16, 16]$, then tile it three times to form a $[384, 3, 16, 16]$ projection. This preserves the spatial convolution structure while distributing spectral information uniformly across RGB channels; all subsequent transformer weights remain unchanged.
   
    \item \textbf{AlphaEarth:} This model entirely bypasses local GPU processing constraints by providing precomputed 64-dimensional multi-sensor pixel embeddings processed natively within Google Earth Engine, serving as an efficient baseline for resource-constrained large-scale deployments.
\end{itemize}

\subsection{Implementation Details}
To isolate the effect of the visual encoder, all competing foundation model configurations are mapped to an identical downstream WeDAN generative framework. The underlying GraphTransformer neural engine is configured with 4 multi-head attention layers, incorporating a fixed hidden layer dimensionality of 32. The forward corruption pipeline applies a parametric cosine noise schedule to step-by-step inject variance across $T = 250$ discrete diffusion timesteps. The complete structural hyperparameter configuration utilized across all model variations is detailed in Table~\ref{tab:hyperparameters}.

\begin{table}[h]
\centering
\small
\caption{Downstream GraphTransformer and Diffusion Network Hyperparameter Specifications}
\label{tab:hyperparameters}
\begin{tabular}{lc}
\hline
\textbf{Hyperparameter} & \textbf{Value} \\ \hline
GraphTransformer Layers & 4 \\
Attention Head Count ($K$) & 4 \\
Hidden Layer Dimension & 32 \\
Total Diffusion Training Steps ($T$) & 250 \\
Noise Schedule & Cosine Schedule \\
Primary Learning Rate & $1 \times 10^{-3}$ \\
Optimization Algorithm & AdamW \\
Batch Size & 1 \\
Dropout Regularization Rate & 0 \\
DDIM Sampling Steps & 25 \\
Monte Carlo Samples per City & 50 \\ \hline
\end{tabular}
\end{table}

During the inference phase, the iterative flow generation process is completed via deterministic DDIM sampling, traversing a compressed 25-step trajectory. To minimize stochastic variance and prevent single-run anomalies from distorting performance tracking, the final generated flow adjacency matrix for each independent city is calculated as the mean over 50 independent DDIM sampling runs. All model iterations are trained, validated, and evaluated across 5 random seeds $\{2024, 2025, 2026, 2027, 2028\}$. Because all encoders are evaluated on an identical set of seeds, splits, and target regions, statistical differences are validated using two-tailed \emph{paired} $t$-tests over the five matched seeds, with Holm--Bonferroni correction across the three reported comparisons.

\section{Results and Analysis} 

Table~\ref{tab:main_results} consolidates performance across all four encoders, averaged over five random seeds and all census-noise levels ($\eta \in \{0, 0.25, 0.5, 0.75, 1\}$). Tables~\ref{tab:eta_us}--\ref{tab:eta_global} provide a granular breakdown of spatial allocation alignment and absolute error metrics across five discrete census availability tiers ($\eta \in \{0, 0.25, 0.5, 0.75, 1\}$).

\begin{table*}[t]
\centering
\small
\caption{Mean ($\pm$std) performance across 5 seeds and all $\eta$ levels. CPC ($\uparrow$) represents the primary metric for spatial allocation alignment; RMSE, NRMSE, and MAE are lower-is-better ($\downarrow$). \textsuperscript{$\dagger$}SatCLIP Global metrics are evaluated over $n{=}4$ seeds. \textsuperscript{$\ddagger$}Global NRMSE exhibits high scale-sensitivity; CPC serves as the primary metric.} 
\label{tab:main_results}
\begin{tabular}{lcccccccccccc} 
\toprule 
& \multicolumn{4}{c}{\textbf{US (In-Distribution)}} 
& \multicolumn{4}{c}{\textbf{UK (Zero-Shot Transfer)}} 
& \multicolumn{4}{c}{\textbf{Global (Zero-Shot Transfer)}} \\
\cmidrule(lr){2-5}\cmidrule(lr){6-9}\cmidrule(lr){10-13} 
\textbf{Model Paradigm} & CPC & RMSE & NRMSE & MAE & CPC & RMSE & NRMSE & MAE & CPC & RMSE & NRMSE & MAE \\ 
\midrule 
RemoteCLIP 
  & \textbf{0.598\,{\small$\pm$0.008}} & \textbf{68.1} & \textbf{0.955} & \textbf{39.7} 
  & 0.349\,{\small$\pm$0.058} & 104.5 & 2.08 & 71.5 
  & \textbf{0.122\,{\small$\pm$0.014}} & 632 & 1927\textsuperscript{$\ddagger$} & 205.8 \\ 
AlphaEarth 
  & 0.568\,{\small$\pm$0.019} & 71.1 & 0.967 & 41.8 
  & 0.455\,{\small$\pm$0.118} & \textbf{79.7} & \textbf{1.55} & \textbf{54.1} 
  & 0.111\,{\small$\pm$0.017} & \textbf{613} & \textbf{488} & \textbf{201.3} \\
DINOv3 
  & 0.509\,{\small$\pm$0.009} & 77.4 & 1.037 & 43.7 
  & 0.329\,{\small$\pm$0.060} & 342.1 & 8.26 & 199.9
  & 0.022\,{\small$\pm$0.006} & 1083 & 10054\textsuperscript{$\ddagger$} & 580.7 \\
SatCLIP\textsuperscript{$\dagger$} 
  & 0.555\,{\small$\pm$0.027} & 72.8 & 0.992 & 43.3 
  & \textbf{0.495\,{\small$\pm$0.032}} & 83.4 & 1.60 & 59.5 
  & 0.104\,{\small$\pm$0.024} & 666 & 5045\textsuperscript{$\ddagger$} & 214.0 \\ 
\bottomrule 
\end{tabular} 
\end{table*} 

\begin{figure*}[!tbp]
  \centering
  \includegraphics[width=\textwidth]{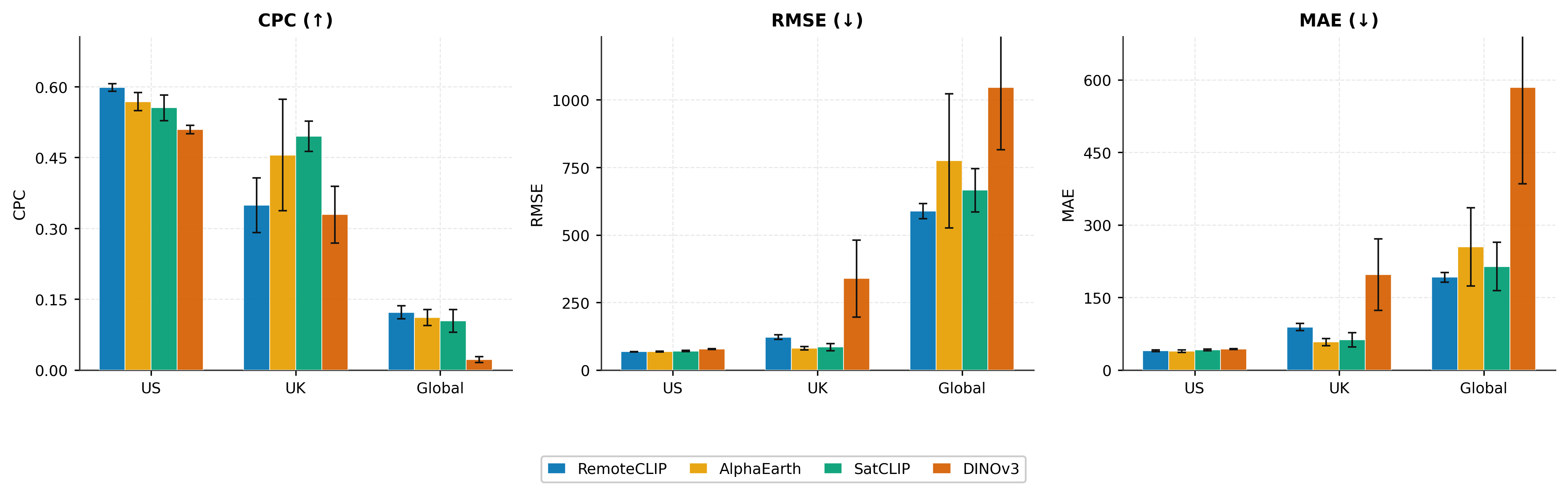}
  \caption{Mean CPC, RMSE, and MAE per region across all $\eta$ levels (mean $\pm$ std, 5 seeds), consistent with Table~\ref{tab:main_results}. RemoteCLIP leads in-distribution (US); SatCLIP and AlphaEarth dominate zero-shot UK transfer; all encoders converge on Global cities. DINOv3 collapses to near-zero performance globally ($\text{CPC} = 0.022$).}
  \label{fig:encoder_comparison}
\end{figure*}

\begin{figure*}[t]
  \centering
  \includegraphics[width=\textwidth]{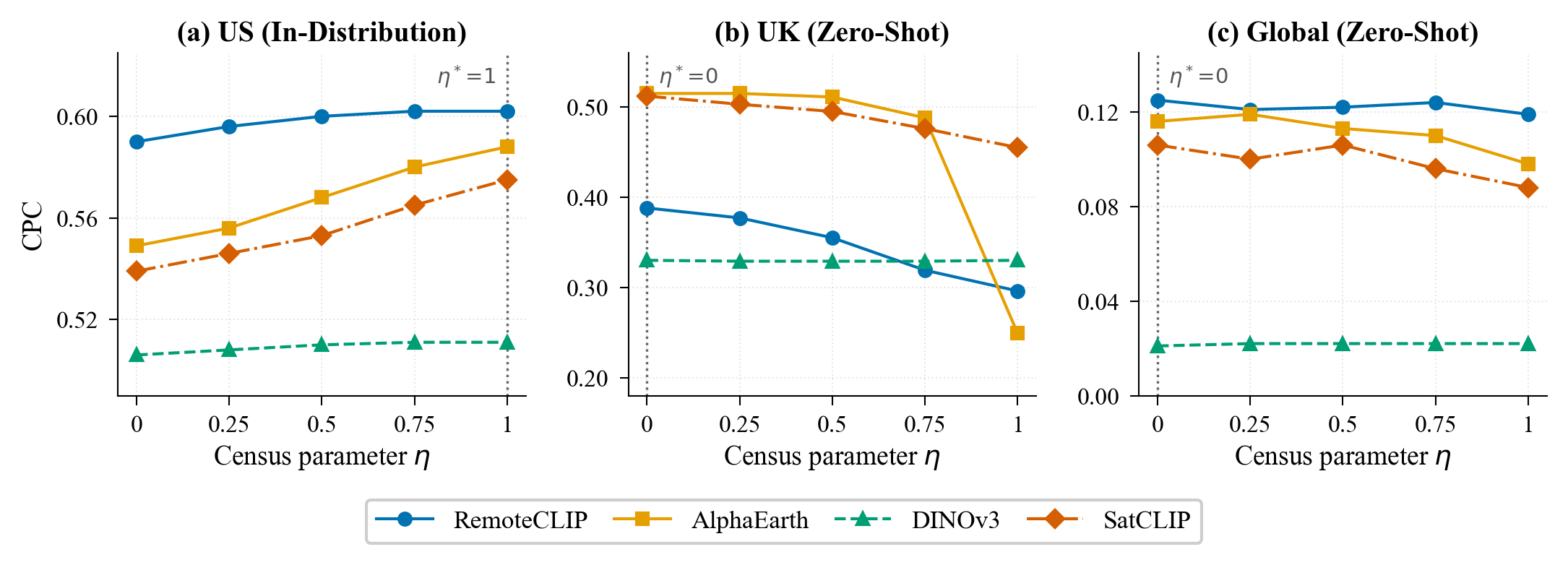}
  \caption{CPC as a function of census noise parameter $\eta$ across all four encoders and three evaluation benchmarks (mean over 5 seeds). The dashed vertical line marks the primary evaluation point for each dataset ($\eta{=}1$ for US; $\eta{=}0$ for UK and Global). US performance increases monotonically with census availability, whereas UK and Global performance degrades as $\eta \to 1$ due to the introduction of out-of-distribution zero-valued census vectors.}
  \label{fig:cpc_eta}
\end{figure*}

\subsection{US In-Distribution Performance}
As shown in Table~\ref{tab:main_results}, RemoteCLIP achieves the highest overall mean CPC ($0.598 \pm 0.008$). When isolating the primary clean evaluation point ($\eta = 1$, full census available), this performance reaches $0.602 \pm 0.009$, recovering 94.9\% of the full-census WeDAN baseline (0.634) reported by GlODGen~\cite{glodgen}. AlphaEarth and SatCLIP trail closely within the macro averages, while DINOv3 lags significantly with an overall mean CPC of $0.509 \pm 0.009$ ($p < 0.001$, paired $t$-test vs.\ RemoteCLIP). 

This gap is particularly notable given DINOv3's 493-million-image pretraining corpus: raw scale does not compensate for misaligned objectives. The self-distillation objective specializes in micro-scale surface geometry and texture discrimination, whereas commuting OD generation requires semantic features of human mobility patterns, the precise signal that language-aligned contrastive supervision encodes. A plausible explanation, which we do not test directly, is an \emph{aggregation mismatch} between DINOv3's pretraining objective and the zone-level prediction task. DINOv3's teacher-student self-distillation is optimized to discriminate between individual image patches at fine spatial scales. Each US county contains 50--200 H3 resolution-9 tiles, which are independently encoded and mean-pooled into a single zone embedding. Mean-pooling of patch-discriminative features destroys the very discriminability the encoder was trained to produce: if each tile captures locally distinctive surface geometry (a parking lot, a rooftop, a tree canopy), but the encoder has not learned to associate these patterns with zone-level commuting character, the pooled embedding carries near-zero relevant signal. RemoteCLIP's contrastive language supervision, by contrast, produces \emph{scene-level} embeddings aligned to high-level semantic categories (e.g., "dense urban commercial area", "low-density suburban residential") that remain meaningfully discriminative after mean-pooling because they already operate at the semantic granularity required for zone-level conditioning.

All four encoders generally exhibit a consistent increase in CPC as $\eta$ increases from 0 to 1 (Figure~\ref{fig:cpc_eta}), confirming that census features provide additive conditioning value when their distribution remains in-domain.

\subsection{UK Zero-Shot Transfer}
The ranking inverts sharply under zero-shot cross-continental transfer. At the declared UK operating point ($\eta = 0$, local census entirely unavailable), AlphaEarth reaches $0.515 \pm 0.015$ and SatCLIP $0.513 \pm 0.034$, a difference that is not statistically significant ($p = 0.91$, paired $t$-test). RemoteCLIP attains only $0.388$ and DINOv3 $0.330$. Averaged across all $\eta$ levels (Table~\ref{tab:main_results}), SatCLIP and AlphaEarth exchange places ($0.495 \pm 0.032$ vs.\ $0.455 \pm 0.118$), reflecting AlphaEarth's sharper degradation as $\eta \to 1$; RemoteCLIP and DINOv3 remain far behind at $0.349 \pm 0.058$ and $0.329 \pm 0.060$. The clear performance advantage of AlphaEarth and SatCLIP over RemoteCLIP constitutes a central empirical finding of this work: \emph{geographically grounded encoders generalize better across continents than language-aligned features}.

One explanation consistent with these results is distributional overfitting. RemoteCLIP's language supervision encodes human mobility pattern semantics specific to North American urban morphology: low-density suburban sprawl, freeway-anchored commercial strips, which do not translate as effectively to European land-use configurations. Conversely, AlphaEarth's multi-sensor physical embeddings (spectral indices, radar backscatter, elevation) and SatCLIP's coordinate-contrastive pretraining may encode geophysical and geospatial priors that remain continuous across continental boundaries.

Within each independent seed, DINOv3's performance is conspicuously insensitive to variation in $\eta$ on UK data (Figure~\ref{fig:cpc_eta}), despite a high cross-seed variance that results in an aggregate score of $0.329 \pm 0.060$. This behavioral signature reveals that DINOv3's geometric features carry minimal mobility signal when deployed out-of-distribution: the model's pathway through the census conditioning channel contributes near-zero marginal information, making the network unresponsive to changes in demographic inputs.

\subsection{Global Zero-Shot Generalization}
Under the most distant transfer setting (US $\to$ Global), all encoders experience significant performance degradation. RemoteCLIP achieves the highest Global CPC ($0.122 \pm 0.014$), with AlphaEarth ($0.111 \pm 0.017$) and SatCLIP ($0.104 \pm 0.024$) following in close proximity. DINOv3 yields only $0.022 \pm 0.006$, confirming that texture-specialized self-distillation features provide little viable cross-continental transfer signal. For reference, classical Gravity models report CPC $\approx 0.32$ on comparable US commuting data~\cite{gravity}. Every satellite-conditioned configuration evaluated here falls far below that on the global benchmark, placing global performance beneath even a classical spatial-interaction prior.

The near-convergence of RemoteCLIP, AlphaEarth, and SatCLIP on the Global benchmark ($\Delta\text{CPC} < 0.025$) suggests that at this extreme transfer distance, encoder choice becomes secondary to the fundamental distribution shift between training domains and morphologically heterogeneous global metropolises. Reported NRMSE values for Global are highly unstable across seeds (e.g., RemoteCLIP: 1927) due to normalization against city-level mean flows, which vary by orders of magnitude across the 14 target cities; CPC should be treated as the sole reliable metric on this benchmark.

Four compounding factors explain this systematic global collapse. \textbf{(i) Morphological diversity:} the 14 evaluation cities span radically distinct urban forms: Tokyo's transit-oriented polycentric structure, Rio de Janeiro's favela-formal city duality, and peri-urban sub-Saharan settlement patterns share no common visual or demographic signature with US county-level commuting distributions. \textbf{(ii) Scale mismatch:} the training distribution contains cities with approximately 30-200 zones; London alone contains 932 zones, roughly five times the largest training city, placing its graph structure firmly out of distribution for the GraphTransformer's attention layers. \textbf{(iii) OD data heterogeneity:} ground-truth flows for global cities are aggregated from heterogeneous sources, household travel surveys, call detail records, and mobile signaling registries, each introducing distinct noise floors and systematic collection biases that the CPC metric conflates with model error. \textbf{(iv) Zero-census conditioning:} all global evaluation operates at $\eta{=}0$, already the weaker conditioning regime; combined with the domain shift in satellite appearance, neither modality provides sufficient signal to anchor the diffusion process to local mobility patterns. Taken together, these factors indicate that the global performance gap is not an encoder-selection problem but a fundamental distributional mismatch requiring either local fine-tuning data or geographically diverse training corpora.

\subsection{Census Noise Sensitivity and Operational Implications} 
The $\eta$-sensitivity patterns across datasets empirically validate the architectural interpretation developed in Section 3. For US data, all encoders benefit from increasing $\eta$ (more census signal). For UK and Global data, the opposite holds: RemoteCLIP and AlphaEarth CPC values degrade as $\eta \to 1$ because supplying zero-valued census vectors at $\eta = 1$ introduces an out-of-distribution input that the network was not exposed to during training (Figure~\ref{fig:cpc_eta}). Practitioners deploying these models in census-absent regions must evaluate strictly at $\eta = 0$ to avoid silent performance degradation. This finding has direct operational consequences: tools and frameworks built on WeDAN should expose $\eta$ as a deployment-time parameter rather than fixing it at training time, allowing practitioners to select the appropriate regime based on local census availability. This heterogeneity carries a direct deployment implication. Although AlphaEarth attains the best UK transfer at $\eta = 0$, it is also the most sensitive to the census channel: its UK CPC falls from 0.515 at $\eta = 0$ to 0.249 at $\eta = 1$, whereas SatCLIP degrades only from 0.512 to 0.455 over the same range (Table~\ref{tab:eta_uk}). A practitioner holding partial or unreliable census data should therefore prefer SatCLIP, whose accuracy is statistically indistinguishable from AlphaEarth's at $\eta = 0$ but far more stable as census information is introduced; AlphaEarth is the better choice only when the census channel is guaranteed to be absent.

\subsection{Practical Encoder Selection}
Table~\ref{tab:compute} summarizes the computational profile of each encoder. AlphaEarth is the only configuration requiring no local GPU: all feature extraction runs natively within Google Earth Engine (GEE), making it the default choice for resource-constrained or cloud-first deployments. SatCLIP's ViT-S/16 backbone incurs roughly one-third the VRAM of the ViT-L models, enabling extraction on consumer-grade hardware. RemoteCLIP and DINOv3 share identical backbone size (ViT-L) and extraction cost; the performance gap between them is attributable entirely to pretraining objective rather than model capacity.

\begin{table}[tbp]
\centering
\small
\caption{Computational profiles of evaluated encoders. GEE = Google Earth Engine cloud extraction; H200 = NVIDIA H200 GPU.}
\label{tab:compute}
\begin{tabular}{lcccc}
\hline
\textbf{Encoder} & \textbf{Backbone} & \textbf{Dim} & \textbf{Hardware} & \textbf{Params} \\ \hline
RemoteCLIP & ViT-L/14 & 1024 & H200 GPU & 307M \\
DINOv3     & ViT-L/16 & 1024 & H200 GPU & 307M \\
SatCLIP    & ViT-S/16 &  384 & H200 GPU      &  22M \\
AlphaEarth & GEE fusion &   64 & GEE (CPU) & --- \\ \hline
\end{tabular}
\end{table}

Table~\ref{tab:recommendation} distills the empirical findings into deployment recommendations. No single encoder dominates across all settings; the optimal choice depends on deployment region, census data availability, and hardware constraints.

\begin{table}[tbp]
\centering
\small
\caption{Encoder selection guide by deployment scenario.}
\label{tab:recommendation}
\begin{tabular}{p{2.8cm}p{1.6cm}p{2.6cm}}
\hline
\textbf{Scenario} & \textbf{Encoder} & \textbf{Rationale} \\ \hline
In-dist., census available            & RemoteCLIP  & Highest US CPC (0.602) \\
Zero-shot, Europe                     & AlphaEarth\slash SatCLIP & Tied at $\eta{=}0$ (0.515 vs.\ 0.513, n.s.) \\
Zero-shot, no GPU                     & AlphaEarth  & GEE cloud extraction \\
Minimal embedding dim                 & AlphaEarth  & 64-dim; low memory \\
Diverse global cities                 & Any except DINOv3$^*$ & Gap $<$0.025 CPC \\ \hline
\multicolumn{3}{l}{\small $^*$These three converge near CPC $\approx$ 0.11 globally; DINOv3 reaches only 0.022.}
\end{tabular}
\end{table}

\section{Limitations and Future Work}
\subsection{Limitations}
\textbf{Evaluation Limitations.} Several structural constraints bound the scope of the present evaluation. First, all evaluated encoders operate on static annual satellite composites. Commuting flows, however, are inherently dynamic: rush-hour congestion patterns, seasonal employment shifts, and post-pandemic transit adjustments introduce temporal variance that no single static visual composite can capture.

Second, our 80/10/10 train/valid/test split reduces training data by approximately 190 cities relative to the GlODGen baseline (90/10), introducing a systematic 2--3\% CPC deficit. Beyond this methodological gap, deploying models trained on North American urban morphology to UK LADs and global cities introduces domain shifts that zero-shot visual embeddings only partially bridge, evidenced by a 37\% CPC drop from the US to the UK (e.g., 0.602 $\to$ 0.388 for RemoteCLIP).

Third, while SatCLIP demonstrated robust cross-continental resilience via coordinate-contrastive alignment, its architectural reliance on standard optical patches limits its ability to fully exploit the rich, multi-spectral geophysical signatures (e.g., radar backscatter, elevation) natively captured by AlphaEarth's multi-sensor fusion.

Finally, the DINOv3 evaluation demonstrates that raw data scale does not guarantee task alignment. Despite a large-scale pretraining corpus, the model's empirical performance remained constrained by its self-distillation objective, which over-indexes on patch-level texture discrimination rather than the zone-level functional land use required for OD flow generation.

\textbf{OD Metric Limitations.} A further limitation concerns evaluation granularity. Our primary metric, CPC, aggregates flow accuracy over all origin-destination pairs equally, masking systematic directional biases—for instance, consistent over-prediction of short-distance intra-zone flows or under-prediction of long-distance commutes. Decomposing CPC by distance decay bin, flow magnitude quantile, or intra- versus inter-zone flows would provide a more diagnostically useful performance profile and expose encoder-specific failure modes that aggregate metrics obscure. Similarly, our Global results are reported as means over 14 cities, which masks substantial per-city variation; a per-city breakdown would be required to separate morphological effects from reference-data quality.

\textbf{Backbone Architectural Constraints and Alternatives.} A final limitation concerns the generative backbone itself. All experiments fix WeDAN as the graph diffusion architecture, varying only the visual encoder. WeDAN's GraphTransformer operates with a batch size of 1 (one city per gradient step), constraining cross-city structural learning. Its full-attention mechanism scales quadratically with node count, restricting deployment to cities with fewer than approximately 1,000 zones. Alternative generative backbones warrant investigation: score-based graph diffusion via stochastic differential equations~\cite{gdss} and variational flow matching formulations~\cite{vfm_graph} offer potentially superior scalability and generalization properties. Disentangling encoder quality from backbone choice remains an open question for future work.

\textbf{Reproducibility Constraints.} 
AlphaEarth embeddings depend on Google Earth Engine asset availability and API versioning, which may change over time. The SatCLIP 13-to-3 channel band weight averaging is a non-standard adaptation; exact reproduction requires the released extraction code. All runs depend on the frozen DenoisingTransformer checkpoint sourced from GlODGen~\cite{glodgen}.% Satellite tile access via Esri World Imagery requires an API token.

\subsection{Future Work}
Several directions emerge directly from our findings.

\textbf{Traffic-encoded node features.} Static satellite composites cannot observe dynamic mobility signals such as peak-hour congestion, transit ridership, or road capacity utilization. Incorporating zone-level traffic indicators, from loop detector networks (TMAS), GTFS transit feeds, or OSM road structural features, as additional node signals would provide employment-density proxies orthogonal to satellite appearance, potentially closing the gap between the $\eta{=}0$ and $\eta{=}1$ operating points.

\textbf{Multi-temporal satellite features.} Extending encoders to multi-temporal composites (quarterly Sentinel-2 stacks, VIIRS nighttime light time series, or Sentinel-1 SAR coherence change) would capture seasonal land-use dynamics invisible to annual composites. Pretrained temporal encoders such as Prithvi-TS could be integrated with minimal architectural modification.

\textbf{Architectural scaling and fine-tuning.} Architecturally, WeDAN relies on a standard full-attention GraphTransformer, which scales quadratically with the number of nodes. Upgrading the backbone with sparse or flash attention mechanisms would enable the model to process significantly larger metropolitan graphs without memory bottlenecks, improving macro-scale flow estimation accuracy. Furthermore, exploring end-to-end fine-tuning paradigms via Low-Rank Adaptation (LoRA)~\cite{lora} could tailor the frozen spatial embeddings explicitly for systemic network flow rather than generic visual similarity.

\textbf{Few-shot adaptation for global cities.} To address the cross-continental split discrepancy, future research should explore few-shot global adaptation. Rather than relying purely on zero-shot transfer, meta-learning approaches (e.g., MAML~\cite{maml} or Reptile~\cite{reptile}) could enable fast adaptation by exposing the diffusion model to a small held-out subset of target-region zones during fine-tuning, re-anchoring layout distributions with minimal target-domain supervision.

\textbf{Uncertainty quantification.} Finally, the generative nature of the diffusion process natively affords uncertainty quantification. Future iterations can leverage the system's existing generative capacity by exposing the per-sample variance across the 50 DDIM draws as calibrated confidence intervals. Reporting this uncertainty around predicted OD flow pairs would allow transport planners to identify unreliable flow estimates prior to downstream analysis.

\section{Conclusion}
We compared four spatial vision foundation models within a fixed generative OD graph diffusion pipeline. Across 1,925 US counties, 325 UK districts, and 14 global cities, three core findings emerge. First, language-supervised representations (RemoteCLIP) capture structural dependencies accurately inside the training domain but exhibit fragile cross-continental generalization due to morphological distribution shifts; RemoteCLIP achieves a leading CPC of 0.602 on US counties, but drops sharply under transfer conditions. Second, geographical and physical grounding provides superior resilience for zero-shot transfer across continents; AlphaEarth leads UK zero-shot transfer at a peak CPC of 0.515 at $\eta{=}0$, representing a 33\% relative improvement over RemoteCLIP at that operating point. Third, massive pretraining dataset volume alone cannot compensate for task-specific domain alignment, as evidenced by DINOv3's empirical performance limitations and near-zero global CPC. These outcomes suggest that embedding coordinate-contrastive or geophysical structural priors within deep vision architectures is a promising direction for geographically transferable urban mobility generation at scale.

\begin{acks}
We thank the support of the ARC Centre of Excellence for Automated Decision-Making and Society (CE200100005). This research includes computations using the computational cluster Katana supported by Research Technology Services at UNSW Sydney~\cite{katana2010}. Satellite imagery was accessed via Google Earth Engine. US commuting data from LODES (US Census Bureau); UK and global city data from~\cite{glodgen}: UK flows from Office for National Statistics (ONS) registries, global flows from the travel-survey and mobile-signalling sources documented therein. WorldPop population grids provided by the WorldPop Project (University of Southampton). The WeDAN architecture and GlODGen codebase are due to~\cite{glodgen} (Tsinghua University FIB Lab).
\end{acks}

\bibliographystyle{ACM-Reference-Format}
\bibliography{sample-base} 

\clearpage
\appendix
\raggedbottom
\section{Complete Per-Region Results}

\begin{table}[H]
\centering
\renewcommand{\arraystretch}{1.15}
\setlength{\tabcolsep}{11pt}
\caption{US performance metrics by $\eta$ level (mean over seeds).}
\label{tab:eta_us}
\begin{tabular}{llccc}
\hline
\textbf{Encoder} & $\boldsymbol{\eta}$ & \textbf{CPC}$\uparrow$ & \textbf{RMSE}$\downarrow$ & \textbf{MAE}$\downarrow$ \\ \hline
           & $0$    & \textbf{0.590} & \textbf{69.0} & \textbf{40.1} \\
           & $0.25$ & \textbf{0.596} & \textbf{68.2} & \textbf{39.9} \\
RemoteCLIP & $0.5$  & \textbf{0.600} & \textbf{67.9} & \textbf{39.6} \\
           & $0.75$ & \textbf{0.602} & \textbf{67.4} & \textbf{39.4} \\
           & $1$    & \textbf{0.602} & \textbf{67.8} & 39.6 \\ \hline
& $0$    & 0.549 & 73.0 & 42.8 \\
           & $0.25$ & 0.556 & 72.6 & 43.0 \\
AlphaEarth            & $0.5$  & 0.568 & 71.5 & 42.5 \\
           & $0.75$ & 0.580 & 70.1 & 41.4 \\
           & $1$    & 0.588 & 68.2 & \textbf{39.0} \\ \hline
           & $0$    & 0.539 & 74.9 & 44.6 \\
           & $0.25$ & 0.546 & 73.9 & 44.2 \\
SatCLIP    & $0.5$  & 0.552 & 73.3 & 43.8 \\
           & $0.75$ & 0.565 & 71.6 & 42.6 \\
           & $1$    & 0.575 & 70.2 & 41.1 \\ \hline
           & $0$    & 0.506 & 78.3 & 44.3 \\
           & $0.25$ & 0.508 & 77.6 & 44.0 \\
DINOv3     & $0.5$  & 0.510 & 77.5 & 43.8 \\
           & $0.75$ & 0.511 & 77.2 & 43.4 \\
           & $1$    & 0.511 & 76.6 & 42.9 \\ \hline
\end{tabular}
\end{table}

\begin{table}[H]
\centering
\renewcommand{\arraystretch}{1.15}
\setlength{\tabcolsep}{11pt}
\caption{UK performance metrics by $\eta$ level (mean over seeds).}
\label{tab:eta_uk}
\begin{tabular}{llccc}
\hline
\textbf{Encoder} & $\boldsymbol{\eta}$ & \textbf{CPC}$\uparrow$ & \textbf{RMSE}$\downarrow$ & \textbf{MAE}$\downarrow$ \\ \hline
 & $0$    & 0.388 & 121.6 & 88.5 \\
           & $0.25$ & 0.377 & 112.5 & 79.2 \\
RemoteCLIP           & $0.5$  & 0.355 & 102.8 & 69.2 \\
           & $0.75$ & 0.319 &  93.4 & 60.4 \\
           & $1$    & 0.296 &  89.4 & 57.2 \\ \hline
 & $0$    & \textbf{0.515} & \textbf{80.6} & \textbf{57.5} \\
           & $0.25$ & \textbf{0.515} & \textbf{80.7} & \textbf{57.9} \\
AlphaEarth           & $0.5$  & \textbf{0.510} & \textbf{80.2} & \textbf{56.8} \\
           & $0.75$ & \textbf{0.488} & \textbf{78.1} & \textbf{52.2} \\
           & $1$    & 0.249 & \textbf{78.9} & \textbf{46.4} \\ \hline
    & $0$    & 0.512 & 81.4 & 58.1 \\
           & $0.25$ & 0.503 & 84.6 & 62.2 \\
SatCLIP           & $0.5$  & 0.495 & 84.6 & 61.5 \\
           & $0.75$ & 0.476 & 83.9 & 59.2 \\
           & $1$    & \textbf{0.455} & 82.8 & 56.7 \\ \hline
     & $0$    & 0.330 & 338.2 & 197.0 \\
           & $0.25$ & 0.329 & 341.9 & 199.9 \\
DINOv3           & $0.5$  & 0.329 & 342.7 & 200.4 \\
           & $0.75$ & 0.329 & 344.4 & 201.6 \\
           & $1$    & 0.330 & 343.4 & 200.6 \\ \hline
\end{tabular}
\end{table}

\begin{table}[H]
\centering
\renewcommand{\arraystretch}{1.15}
\setlength{\tabcolsep}{11pt}
\caption{Global performance metrics by $\eta$ level (mean over seeds). $^{\dagger}n{=}4$ seeds; see Table~\ref{tab:main_results}.}
\label{tab:eta_global}
\begin{tabular}{llccc}
\hline
\textbf{Encoder} & $\boldsymbol{\eta}$ & \textbf{CPC}$\uparrow$ & \textbf{RMSE}$\downarrow$ & \textbf{MAE}$\downarrow$ \\ \hline
 & $0$    & \textbf{0.125} & \textbf{587.7} & \textbf{191.5} \\
           & $0.25$ & \textbf{0.121} & 690.7 & 225.2 \\
RemoteCLIP           & $0.5$  & \textbf{0.122} & 588.9 & 191.8 \\
           & $0.75$ & \textbf{0.124} & 604.8 & 195.9 \\
           & $1$    & \textbf{0.119} & 689.7 & 224.5 \\ \hline
 & $0$    & 0.116 & 774.6 & 254.2 \\
           & $0.25$ & 0.119 & \textbf{577.7} & \textbf{189.7} \\
AlphaEarth           & $0.5$  & 0.113 & \textbf{573.4} & \textbf{189.8} \\
           & $0.75$ & 0.110 & \textbf{571.9} & \textbf{188.8} \\
           & $1$    & 0.098 & \textbf{566.4} & \textbf{184.1} \\ \hline
 & $0$    & 0.114 & 607.6 & 195.1 \\
           & $0.25$ & 0.106 & 737.6 & 237.5 \\
SatCLIP$^\dagger$           & $0.5$  & 0.112 & 615.9 & 197.2 \\
           & $0.75$ & 0.099 & 621.1 & 199.3 \\
           & $1$    & 0.091 & 748.2 & 241.0 \\ \hline
     & $0$    & 0.021 & 1045.8 & 583.9 \\
           & $0.25$ & 0.022 & 1114.0 & 584.5 \\
DINOv3           & $0.5$  & 0.022 & 1012.4 & 550.3 \\
           & $0.75$ & 0.022 & 1106.3 & 578.9 \\
           & $1$    & 0.022 & 1135.9 & 606.0 \\ \hline
\end{tabular}
\end{table}

\end{document}